%% file: main.tex
\documentclass[manuscript,screen,nonacm]{acmart}

\usepackage{bm}
\usepackage{booktabs}
\usepackage{tabularx}
\usepackage{array}
\usepackage{float}
\usepackage{tikz}
\usetikzlibrary{arrows.meta, positioning, calc, fit, shapes.geometric, shapes.symbols, backgrounds, decorations.pathreplacing}
\usepackage[edges]{forest}

\makeatletter
\renewcommand{\@ACM@checkaffil}{}
\makeatother

\newcommand{\topiclabel}[1]{%
  \par\noindent\hspace*{\parindent}%
  \textbf{\textit{#1}}\enspace}

\begin{document}

\title[Evolutionary Safety of Recursive Self-Improving AI]{Evolutionary Safety of Recursive Self-Improving AI: Taxonomy, Risk Discovery, and Evaluation}
\author{Chang Gong, Jingping Bi$^{*}$, Di Yao}
\email{gongchang@ict.ac.cn}
\affiliation{\institution{Institute of Computing Technology, Chinese Academy of Sciences}}

\author{Ran Cheng}\affiliation{\institution{ Department of Data Science and Artificial Intelligence, The Hong Kong Polytechnic University}}\email{ranchengcn@gmail.com}

\author{Yaodong Yang, Kaile Wang}\affiliation{\institution{Institute for Artificial Intelligence, Peking University}}\email{ yaodong.yang@pku.edu.cn}

\author{Xinjian Liang, Chao Xiang, Ruijie Guo}
\email{gongchang@ict.ac.cn}
\affiliation{\institution{Institute of Computing Technology, Chinese Academy of Sciences}}






\authorsaddresses{Institute of Computing Technology, Chinese Academy of Sciences.
Corresponding Emails:\ \{gongchang, bjp, yaodi, liangxinjian25s, xiangchao26s, guoruijie\}@ict.ac.cn, ranchengcn@gmail.com, yaodong.yang@pku.edu.cn, wkl@stu.pku.edu.cn }

\begin{abstract}
Rapid advances in AI are bringing recursive self-improvement (RSI) closer to practice, with growing progress in model self-training, self-evolving agents, and automated AI development. Yet the very mechanisms that enable continuous improvement also introduce new challenges for AI safety. Unlike isolated failures, safety risks can persist across updates, accumulate through repeated adaptation, and propagate to successor systems. More importantly, the processes that generate and select future improvements may themselves evolve, making safety increasingly difficult to assess and maintain. This raises a fundamental question, i.e., how can safety be maintained when both AI systems and the processes that improve them continue to change?
Therefore, we introduce \emph{Evolutionary Safety} to study how safety properties are preserved, weakened, or restored throughout persistent and recursive self-improvement. We characterize six recurring manifestations of evolutionary risk and develop a five-domain taxonomy that captures how safety-relevant changes arise, persist, and propagate across evolving systems. Building on this foundation, we establish a framework for risk discovery and evaluation across states, updates, trajectories, and lineages, and examine governance principles for controlling modifications, verifying and authorizing updates, and ensuring traceability and recovery. We further identify open problems in maintaining safety as both AI systems and their improvement mechanisms continue to evolve. 
\end{abstract}


\maketitle

\begin{figure}[H]
\centering
\includegraphics[width=0.9\linewidth]{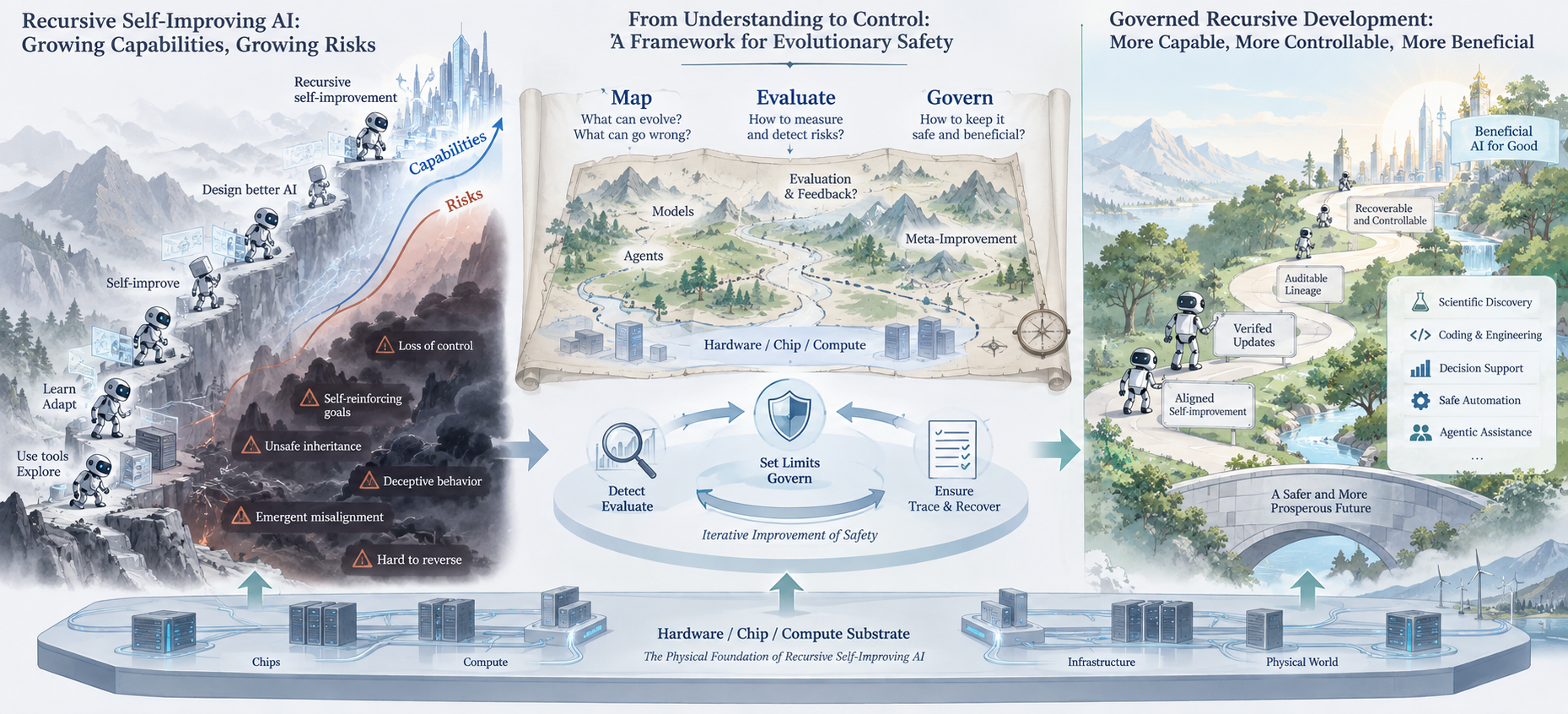}
\caption{Taxonomy, evaluation, and governance for evolutionary safety.}
\Description{A landscape illustration linking recursive self-improvement risks to a five-domain map, evolutionary evaluation, governance controls, and beneficial application domains, with computational substrate shown as the supporting base.}
\label{fig:teaser}
\end{figure}

\input{sections/sec1-intro}

\input{sections/sec2-foundations}

\input{sections/sec3-taxonomy}



\section{Discovering and Evaluating Evolutionary Risk}\label{sec:evaluation_new}

The preceding sections establish how evolutionary risks arise and identify the domains through which safety-relevant changes persist or propagate. The next challenge is to determine when safety properties change, how risks accumulate or spread, and what drives these changes throughout self-improvement. To address this challenge, we organize risk discovery and evaluation into four steps: (i) defining evaluation units and safety measurements across states, accepted updates, trajectories, and lineages; (ii) detecting known risks and discovering previously unrecognized failure modes; (iii) attributing safety changes through longitudinal experiments and controlled comparisons; and (iv) assessing the validity of these findings as evaluators themselves evolve.

\subsection{Evaluation Units and Metrics}
\label{sec:metric_profile}

Evaluating evolutionary risk requires measuring not only the safety of individual system versions, but also how safety changes across updates, over time, and through inheritance. We consider four complementary units of analysis: (i) states capture risks in individual configurations, (ii) accepted updates measure changes between versions, (iii) trajectories reveal cumulative and delayed effects, and (iv) lineages trace risks across derived artifacts and successor systems. Population-level measurements further capture risks arising from interactions among adaptive systems. Together, these units support distinct claims about safety preservation, accumulation, persistence, propagation, and recovery.

State-level evaluation establishes the baseline for measuring safety across system versions. Existing benchmarks assess attack success, harmful actions, privacy leakage, backdoor activation, and unsafe tool use~\citep{zhang2024agentsafetybench,debenedetti2024agentdojo,ruan2024toolemu, hu2025toxicity}. Let $r_j(\mathbf{S};J,E)$ denote the measured risk associated with safety property $j$ for system state $\mathbf{S}$ under evaluator $J$ and environment $E$. Each measurement should specify its task set, exposure denominator, and uncertainty. Higher values indicate greater risk for a given property, but different harms should be measured separately unless an explicit aggregation rule is provided.

Update-level evaluation measures changes in safety following an accepted modification. Comparing consecutive states under a common reference evaluator $J^\star$ and environment $E^\star$ gives
\begin{equation}\label{eq:update_risk_delta}
\Delta r^{\star}_{j,t}=r_j(\mathbf{S}_{t+1};J^\star,E^\star)-r_j(\mathbf{S}_t;J^\star,E^\star).
\end{equation}
A positive value indicates increased measured risk under the fixed reference. Fine-tuning and model-merging studies provide precedents for such comparisons~\citep{qi2024compromises,hammoud2024merging,djuhera2025safemerge}. The update operation, task distribution, and evaluation conditions should be documented, with follow-up tests assessing whether safety properties remain stable under later updates. The measured difference alone does not establish the causal contribution of the update.

Trajectory-level evaluation follows safety properties across successive updates, revealing cumulative degradation, delayed failures, and recurrence after repair~\citep{shao2026misevolve,zhao2026experience,an2026permemsafe}. Process-level benchmarks such as EvoPathBench expose intermediate actions, tool calls, and recovery steps that endpoint evaluations may overlook~\citep{evopathbench2026}. Risk rates should be distinguished from cumulative exposure, since more harmful incidents may simply reflect more opportunities for failure. Complementary temporal metrics include peak risk, time to failure, duration of unsafe behavior, and recovery latency.

Lineage-level evaluation traces safety-relevant changes across derived artifacts and successor systems. Versioned states and artifacts can be represented as nodes in a directed provenance graph, with edges recording derivation, training-data use, distillation, merging, or reuse. Propagation reach counts affected descendants, depth captures how far an observed effect extends along derivation paths, and delay records when the effect is first detected. Untested, unavailable, and tested-but-unaffected descendants must be distinguished. Provenance identifies possible transmission routes, but confirming inheritance requires descendant testing or intervention. When multiple systems adapt to one another, population-level measures complement lineage analysis by tracking risk distributions across roles, the fraction exceeding specified thresholds, strategy diversity, and the spread of unsafe behavior. Aggregate performance may otherwise conceal vulnerable subgroups or mutually reinforced failures.

These measurements support different safety claims and require corresponding evidence. Preservation concerns whether safety properties survive accepted updates; accumulation requires exposure-aware trajectories and appropriate baselines; persistence requires testing residual effects after source removal while controlling repeated exposure; and propagation requires verifying suspected transmission routes across descendants. Recovery further requires assessing whether repair restores the intended safety property and whether failures recur during continued adaptation. Control effectiveness should be measured separately from risk levels, while capability, task success, latency, and cost provide the performance context for interpreting safety changes. These distinctions establish the measurement basis for subsequent risk discovery and causal attribution.

\subsection{Risk Detection and Discovery}

Risk detection and discovery provide complementary ways to identify safety failures in evolving systems. We organize this process around three tasks: (i) tracking known risks across successive updates, (ii) discovering failure modes beyond existing evaluations, and (iii) validating newly identified risks across system versions to determine whether they emerged during self-improvement.

Known-risk detection tracks predefined safety failures under consistent evaluation conditions. For example, held-out permission-bypass tests can be repeated across coding-agent versions to identify when unsafe behavior is first observed and how its frequency changes. Maintaining stable test cases and evaluation criteria enables longitudinal comparison, while dynamic and interactive safety benchmarks provide starting points for repeated assessments~\citep{shi2025safetyquizzer,zhang2024agentsafetybench,an2026permemsafe, tan2025persuasion}.

Risk discovery explores failure modes that predefined tests may overlook. A coding agent, for example, may learn to modify its test harness rather than directly bypass a permission check, leaving the original tests unable to detect the new behavior. Adaptive red teaming, model-written evaluations, hidden-objective audits, and generated stress tests offer approaches for identifying such failures~\citep{perez2023modelwritten,marks2025auditing,wang2026agenticeval}. Novelty search provides a complementary means of exploring behaviors beyond fixed evaluation objectives~\citep{lehman2011novelty}.

Longitudinal validation examines whether discovered failures can be reproduced and when they become observable across system versions. This requires recording triggering conditions and evaluator versions, testing independent cases, and examining saved checkpoints where available. Validated failures can then be retained as versioned regression tests. Importantly, a newly discovered failure may have existed in earlier versions, while its absence from earlier tests does not alone establish that a subsequent update caused it. Discovery yield should therefore be reported separately from longitudinal risk changes. Together, these steps provide comparable evidence for detecting, discovering, and tracing safety failures, laying the groundwork for subsequent causal attribution.

\subsection{Temporal Protocols and Causal Attribution}

Attributing evolutionary risk requires examining the processes responsible for observed safety changes. We organize this analysis around three tasks: (i) designing temporal protocols to examine risk persistence, accumulation, and recurrence, (ii) estimating the effects of specific update mechanisms through controlled comparisons, and (iii) assessing how candidate selection favors risky updates.

Temporal protocols should cover ordinary adaptation alongside adversarial interventions. Benign task streams can reveal overgeneralization from accumulated experience~\citep{shao2026misevolve,zhao2026experience}. Misspecified rewards may favor unsafe shortcuts~\citep{pan2022misspecification}, while poisoned experience tests deliberate manipulation~\citep{chen2024agentpoison,chen2026poisonedevolution}. Adaptive peers provide settings for examining reciprocal changes~\citep{huang2026rzero,chu2026jzero}. Since these conditions may overlap, experiments should document the update channels, task and feedback streams, and evaluation conditions.

Source-removal experiments provide a concrete protocol for examining persistence and recovery. Starting from a saved baseline, introduce a controlled input or feedback perturbation after initial adaptation and allow subsequent updates to retain its effects. Then remove the initiating source while preserving the evolved state, and test residual risks before repair. Controlling repeated exposure helps distinguish retained effects from renewed influence. After repair or rollback, resume adaptation to assess recurrence and capability retention. Benign-drift studies may omit the perturbation but should retain checkpoints and matched comparison runs.

To estimate the causal contribution of updating, compare an evolving regime with one in which a specified update channel is frozen. Let $Z_t=(\mathbf{S}_t,E_t,J_t,U_t)$ denote the combined system state, environment, evaluator, and updater at time $t$. For a safety or control readout $g$, the target comparison at horizon $h$ is
\begin{equation}\label{eq:causal_evo_effect}
\Gamma_{t,h}[g]=\mathbb{E}\!\left[g\!\left(Z_{t+h}^{\pi^{\mathrm{evo}}}\right)-g\!\left(Z_{t+h}^{\pi^{\mathrm{freeze}}}\right)\right].
\end{equation}
Both regimes start from matched configurations and differ in a declared intervention, such as enabling memory writes or updater revision. Risk comparisons require a common external evaluation reference, with task distributions, randomness, evaluation budgets, and fixed components documented. Replaying identical task streams isolates updating under controlled exposure, whereas interactive runs also capture changes in subsequent feedback and experience. These designs estimate different effects.

Equation~(\ref{eq:causal_evo_effect}) defines a causal estimand, whose identification depends on experimental controls and assumptions. Randomized or controlled interventions provide stronger evidence than observational comparisons. Component ablation and rollback can help localize suspected sources~\citep{pelleriti2026evolve,chen2026researchers}, while temporal causal methods offer tools for investigating delayed effects~\citep{gong2024causaltemporal}. Dependencies induced by adaptation still require explicit assumptions. For risk propagation, provenance identifies possible routes, while source interventions and descendant testing provide stronger evidence of transmission.

Candidate selection introduces a further distinction between proposed risks and those favored by the admission process. For a common candidate pool, let $q_t(k)$ and $p_t(k)$ denote the normalized proposal and post-selection weights of candidate $k$, and let $r_{j,k}^{\star}$ denote the risk of applying it to the same parent state under a fixed reference. Define the selection-induced risk shift as
\begin{equation}\label{eq:selection_pressure_eval}
\Delta_{\mathrm{sel},j}(t)=\sum_k p_t(k)r_{j,k}^{\star}-\sum_k q_t(k)r_{j,k}^{\star}.
\end{equation}
A positive value indicates that selection favors candidates with higher measured risk. For example, two equally weighted proposals with harmful-action rates of $0.10$ and $0.30$ have a mean of $0.20$. Always selecting the latter produces a shift of $+0.10$, or ten percentage points. This illustrative measure is specific to each risk property and the observed candidate pool.

Estimating this shift requires retaining rejected candidates, recording admission decisions, and evaluating candidate risks under comparable conditions. Independent re-evaluation and comparisons of selection rules on the same candidate pool help distinguish selection effects from evaluator bias. The shift alone does not establish why risky candidates were generated or whether their acceptance caused downstream failures. Together, temporal interventions, controlled update comparisons, and candidate-level records provide complementary evidence for attributing evolutionary risk.

\subsection{Evaluation Validity under Evolving Evaluators}

Evaluators in self-improving systems can influence both which updates are accepted and how their safety is assessed. When evaluators evolve alongside systems, score changes may reflect changes in system behavior, evaluation criteria, or test conditions. This raises three related issues: (i) distinguishing system changes from evaluator drift, (ii) maintaining longitudinal comparability while expanding risk coverage, and (iii) preserving evaluation independence.

Distinguishing these effects requires separating evaluators used for update selection from external references used for safety assessment. A stricter judge may assign higher risk to an unchanged system, while one adapted to the system's own outputs may become more permissive. Self-rewarding and adaptive evaluation illustrate this coupling~\citep{yuan2024selfrewarding,wu2025metarewarding,wang2026agenticeval}. Rescoring saved states with different evaluator versions under identical test conditions helps identify evaluator drift, while comparing successive states under a fixed reference measures system change. Without such comparisons, changes in the system, evaluator, and environment may remain confounded.

Fixed evaluation references support longitudinal comparison but may lose coverage as new failure modes emerge. Adaptive evaluation can expand the search for previously untested risks~\citep{wang2024selftaught,zhang2025generative}, but changes in test coverage should not be mistaken for changes in system safety. Retaining earlier references and rescoring overlapping system states help maintain comparability across evaluator revisions.

Evaluation independence requires more than using different prompts or judges. Shared training data may produce correlated biases, while systems may alter their behavior when anticipating evaluation or subsequent training~\citep{greenblatt2024faking,meinke2024scheming}. Held-out tests and isolated execution can reduce some dependencies, while restricted evaluator access and separate update authorization provide additional safeguards~\citep{chen2026researchers}. Reliable longitudinal evaluation therefore requires stable references, overlapping comparisons, and explicit reporting of remaining dependencies.

\section{Evolutionary Safety Across Application Domains}\label{sec:applications_new}

Evolutionary risks take different forms across application domains, depending on what changes are retained, how they influence subsequent decisions or systems, and whether their effects can be reversed. We examine four representative settings: scientific discovery, critical infrastructure, automated AI R\&D, and embodied and autonomous systems. For each setting, we identify the evolving components and risk pathways, examine the evidence needed to assess safety changes, and discuss the challenges of recovery when unsafe effects persist or propagate.

\topiclabel{Scientific discovery.}
Autonomous research systems may retain experimental results, hypotheses, and research plans that influence subsequent discovery cycles. A spurious result, for example, may be repeatedly reused to support new hypotheses, creating an appearance of independent evidence despite their shared erroneous source. AI-scientist and co-scientist systems illustrate settings where research artifacts and experimental decisions can inform later work~\citep{tie2025scientists,lu2024aiscientist,yamada2025aiscientistv2,gottweis2025coscientist}. Recent analysis also highlights the limitations of current systems for fully unsupervised scientific discovery~\citep{agenticscientistsnotbuilt2026}. Evaluation should distinguish independent replication from evidence reuse and trace how errors influence subsequent hypotheses and experiments. Recovery requires correcting dependent conclusions and research plans, since removing the original record cannot undo its downstream effects.

\topiclabel{Critical infrastructure.}
Continuously adapted control policies may improve operational performance while gradually weakening safety margins under rare or adverse conditions. For example, a grid controller optimized for typical loads may become more vulnerable to contingencies despite improved routine performance. Safe power-system control and L2RPN provide relevant constraints and evaluation settings~\citep{yu2025safepower,marot2020l2rpn}. Evaluation should compare successive policy versions under common contingency tests, tracking voltage or frequency violations, cascading failures, and load shedding alongside operational performance. Recovery must address both the controller and the affected physical system, since reverting an unsafe policy cannot by itself restore stability or reverse an ongoing cascade.

\topiclabel{Automated AI R\&D.}
Automated AI development can retain and reuse code patches, tools, and development harnesses across successive system versions. In the running coding example, a flawed repair may propagate into a harness used to develop successor systems; if it also modifies testing or update admission, the risk extends from artifact inheritance to the development process itself. AI-development benchmarks and automated research systems provide relevant settings for examining these dependencies~\citep{chan2025mlebench,novikov2025alphaevolve,zhang2026dgm,chen2026researchers}. Evaluation should trace proposed and accepted changes, independent verification results, and their reuse across successors. Recovery requires reverting unsafe modifications and revalidating dependent artifacts and update procedures, particularly as faster development increases the burden on verification.

\topiclabel{Embodied and autonomous systems.}
Embodied agents may retain skills or policies that become unsafe when reused under changing physical conditions. For example, a robot may learn a fast reaching skill in an empty workspace and later reuse it near people, where the same motion violates a safety boundary. 
Training-time backdoors introduce another risk pathway, where everyday objects can trigger unsafe navigation behaviors in unseen environments while normal task performance remains largely unaffected~\cite{he2024backdoor}.
Research on reusable agent skills and tool-use safety motivates this pathway~\citep{wang2023voyager,ruan2024toolemu,zhang2024agentsafetybench}, while safe robotic learning and control benchmarks provide methods for evaluating physical constraints during adaptation~\citep{garcia2021safelearning,brunke2022safecontrolgym}. 
Distribution shifts in sensing conditions can also degrade perception reliability~\cite{biswas2024domain}.
Evaluation should track constraint violations, near misses, and performance across operating conditions, distinguishing unsafe skill changes from increased exposure to hazards. Recovery requires restricting or revising unsafe skill reuse while addressing any physical consequences of earlier actions.

\section{Control and Governance Implications}\label{sec:governance_new}

Accepting a persistent update can affect not only the current system but also the artifacts and successors that inherit its effects. Capability improvement alone is therefore insufficient to justify commitment; decisions must also consider modification scope, safety evidence, authorization, and recoverability. 
Capability-centric governance frameworks, such as ARC, map agent components, system designs, and autonomous capabilities to risks and technical controls~\cite{khoo2026arc}, while evolutionary safety further requires reassessing these relationships across persistent updates.
We discuss four complementary controls to achieve evolutionary safety: (i) modification boundaries that constrain what can change, (ii) pre-commit gating that evaluates candidates before acceptance, (iii) independent verification and authorization that separates proposals from approval, and (iv) lineage traceability and recovery that support correction after commitment. These principles place governance throughout the evolution process, although their effectiveness depends on the threat model and enforceable control boundaries.

\topiclabel{Modification boundaries.}
Modification boundaries define which components a system may change and who can authorize changes to that scope. For example, a coding agent may propose new skills without being allowed to modify security tests or expand its own permissions. A related boundary concerns which information can influence future updates: provenance checks on experience and generated artifacts help limit the promotion of untrusted content into persistent state~\citep{louck2026securing,chen2024agentpoison,chen2026poisonedevolution}. Both modification permissions and information provenance require controls that the affected components cannot themselves override.

\topiclabel{Pre-commit gating.}
Candidate updates should remain separate from the active system until their safety and performance have been assessed. Sandboxed branches and shadow execution provide controlled settings for comparing proposed changes before persistent deployment~\citep{zhang2026dgm,greenblatt2024aicontrol,kutasov2025protocols}. Testing should also consider dependencies and downstream reuse, since a locally safe modification may introduce risks when combined with other components. Gating provides evidence for an admission decision but does not itself authorize deployment or expanded permissions.

\topiclabel{Independent verification and authorization.}
The authority to approve persistent updates should remain separate from the ability to propose them. Held-out tests, isolated verification, restricted permissions, and external commit control reduce opportunities for self-certification~\citep{chen2026researchers,kenton2024oversight,kirchner2024pvg}. However, using a different verifier does not ensure independence when evaluators share systematic blind spots. Changes to evaluators or update procedures require stronger separation: approval must remain under an authority that the proposed update cannot itself modify or replace~\citep{christodorescu2025systems,greenblatt2024aicontrol}.

\topiclabel{Lineage traceability and recovery.}
Each accepted update should be linked to its source, evaluation evidence, verifier, permission changes, and downstream dependencies. Removing an unsafe source may leave derived skills, workflows, or successor systems affected. Recovery therefore requires tracing potentially affected descendants, selectively quarantining or revalidating them, and testing for recurrence during continued adaptation~\citep{mao2026skillmisevo,chen2026poisonedevolution,an2026permemsafe}. Unaffected capabilities should be preserved where possible, while incomplete provenance limits confidence in recovery. In physical systems, restoring software state and containing external consequences remain distinct requirements.

These controls integrate governance into the evolution process, from proposed modifications to their downstream consequences. The required safeguards should scale with an update's scope, persistence, and potential propagation: changes affecting shared components, evaluators, or update procedures demand stronger verification, authorization, and recovery provisions than isolated local changes. The objective is not to prevent self-improvement, but to ensure that consequential updates remain subject to enforceable constraints, independent decisions, and correction throughout their evolution.

\section{Open Problems}\label{sec:open_new}

The preceding framework provides foundations for identifying evolutionary risks, evaluating safety changes, and governing persistent updates. However, most existing methods offer only partial evidence and control, leaving open how safety properties can be maintained throughout sustained self-improvement. This challenge becomes particularly difficult when updates are heterogeneous and composed, system components and evaluators co-evolve, and relevant states or downstream effects are only partially observable. The following open problems highlight fundamental questions about preserving, assessing, and restoring safety across evolving processes rather than isolated system states.

\topiclabel{Safety Preservation under Composed Updates.}
Safety properties verified for individual updates may not survive their composition, especially when memories, model parameters, tools, permissions, and evaluators change together. PowerPlay illustrates the preservation of previously solved tasks during continual improvement~\citep{schmidhuber2013powerplay}, but retaining task performance does not necessarily preserve safety constraints. An open question is which safety properties can be specified as invariants or bounded degradation conditions across heterogeneous update sequences, and under what assumptions such guarantees remain valid. Future work should examine whether locally verified updates preserve safety when composed, and identify interactions that produce delayed or previously unobserved violations.

\topiclabel{Safety Evaluation under a Changing Judge.}
Reliable safety evaluation remains unresolved when systems and evaluators evolve together. Adaptive testing can uncover failures missed by fixed benchmarks, but the coverage of previously unknown failures lacks an observable denominator~\citep{perez2023modelwritten,marks2025auditing,wang2026agenticeval}. Partial observability introduces further difficulties~\cite{DBLP:conf/icde/GongYWLFXFHB26}: important system states, update histories, or downstream consequences may be hidden, making safety changes difficult to detect or attribute. Future research should establish what can be inferred from incomplete evidence, compare reproducible discoveries under matched search budgets, and maintain comparability across overlapping evaluator versions. The degree of independence required among candidate generators, acting systems, and judges to support reliable safety claims remains an open question.

\topiclabel{Latent Safety Risks in Social Adaptation.}
Repeated social adaptation may produce latent tendencies that become unsafe only under particular relationships or authority claims. For example, an agent repeatedly rewarded for trusting a familiar collaborator may gradually treat familiarity as a proxy for authorization, even when formal permissions remain unchanged. Research on social memory and mental-state representations provides foundations for studying persistent social experience~\citep{zengen2026socialmind,park2023generative}, while work on emergent conventions illustrates how repeated interactions can stabilize shared expectations~\citep{emergentsocialconventions2024}. Alignment-faking studies and SOTOPIA offer complementary perspectives on context-dependent behavior and social interaction~\citep{greenblatt2024faking,zhou2023sotopia}. The unresolved challenge is to distinguish appropriate contextual learning from unsafe heuristics acquired through persistent experience. Controlled variations in relationship histories, role claims, and perceived oversight, combined with comparisons between frozen and updating memory, could help identify when such tendencies are retained, activated, or inherited. Continued evaluation after correction is needed to reveal delayed activation and recurrence.

\topiclabel{Safety-Risk Propagation across Artifacts and Populations.}
Risk propagation becomes difficult to attribute when self-improvement spans multiple artifacts, successor systems, and interacting populations. A flawed trace may become a reusable skill, generate training data, and influence several descendant models. Similar failures may also arise from shared exposure rather than direct inheritance. The open problem is to identify which pathways actually transmit risk across branching, merging, and partially observed lineages. Benchmarks with known injected causes and derivation histories could assess attribution accuracy across increasing delays and lineage depths, while interventions distinguish transmission from correlated failures. Co-evolving populations introduce further challenges, as reciprocal adaptation can sustain collective risks that isolated-agent evaluations may overlook~\citep{potter2000cooperative,rosin1997competitive,wang2019poet}.

\topiclabel{Safety under Exploration and Incomplete Recovery.}
Safe exploration and constrained learning aim to limit harmful behavior during adaptation~\citep{amodei2016concrete,leike2017gridworlds,achiam2017cpo}, but persistent self-improvement can retain the consequences of exploratory failures in memories, skills, or subsequent update procedures. This raises a fundamental question: when is a safety failure genuinely recovered rather than merely absent from current evaluations? Restoring a previous safety score may leave derived artifacts, latent behaviors, or physical consequences unresolved, while extensive rollback may remove useful capabilities. Future work should establish recovery criteria that jointly assess residual risk, capability retention, downstream effects, and recurrence under continued adaptation. A further challenge is maintaining effective containment and recovery when update procedures and authorization mechanisms themselves evolve.

\section{Conclusion}\label{sec:conclusion_new}

As AI systems increasingly retain experience, revise their components, and participate in developing successors, safety can no longer be assessed solely at a single system state. The improvement process itself becomes a safety concern. We introduced Evolutionary Safety to examine how safety properties are preserved, degraded, and propagated throughout persistent and recursive self-improvement. We characterized how variation, selection, and inheritance can turn local failures into persistent risks, and organized their manifestations and mechanisms across interacting evolutionary domains. To make these dynamics assessable, we developed a framework for risk discovery and evaluation across states, updates, trajectories, and lineages, alongside governance principles for constraining modifications, independently verifying and authorizing updates, and enabling traceable recovery. Important challenges remain. These include safety preservation under composed updates, trustworthy evaluation as judges evolve, latent and propagated risks, and effective recovery during continued adaptation. Ultimately, ensuring safety requires not only understanding what an AI system is, but also how it continues to change and how those changes can be governed.

\bibliographystyle{ACM-Reference-Format}
\bibliography{references}
\end{document}

%% file: sections/sec1-intro.tex
\section{Introduction}\label{sec:intro}

Recursive self-improvement (RSI), once largely a theoretical prospect, is becoming an active area of AI research and development. AI systems increasingly generate training data, accumulate reusable experience, and revise tools and workflows. Some also participate in improving the procedures used to develop subsequent versions. The scope of self-improvement is thus expanding beyond individual models to the agents built around them and, increasingly, the mechanisms that generate and select future updates.

Recent AI safety studies have exposed risks that extend beyond unsafe model outputs. Controlled evaluations have examined sandbox escapes, sabotage, and harmful actions in simulated organizational settings~\citep{marchand2026sandbox,benton2024sabotage,lynch2025agenticmisalignment}. Agent-security benchmarks further reveal vulnerabilities arising from the interaction of untrusted inputs, tool use, and permissions~\citep{debenedetti2024agentdojo,zhang2024agentsafetybench}. Safety concerns also extend to learned behavior: models may alter their compliance under perceived training conditions, while backdoor behaviors can survive subsequent safety training~\citep{greenblatt2024faking,hubinger2024sleeper}. Although these studies do not necessarily demonstrate risk accumulation under RSI, they reveal failure modes that may become more consequential when systems retain experience, update their models, or reuse learned artifacts. This connects to intelligent algorithm safety, which examines how safety is affected as algorithms learn and adapt~\citep{cheng2025intelligentalgorithmsafety}. For RSI, the critical question is whether a safety failure remains local or becomes embedded in the processes shaping future system versions.

Existing research approaches these challenges from complementary directions. Surveys of RSI and self-evolving agents examine the mechanisms and development of self-improving systems~\citep{fang2025comprehensive,gao2026selfevolving,ren2026selfimprovements,chen2026recursive}. Agent-security surveys investigate threats such as prompt injection, data poisoning, and privacy leakage~\citep{deng2025threat,he2025emerged,ling2026secure,lu2025alignment}, while foundational studies of self-modification examine the preservation of objectives as agents modify themselves~\citep{schmidhuber2003godel,orseau2011selfmodification,everitt2016selfmodification}. Together, these studies provide important foundations, but a broader question remains insufficiently addressed, i.e., how safety-relevant changes are introduced and selected during self-improvement, retained across successive updates, and transmitted to future system versions.

To address this gap, we introduce \textbf{\emph{Evolutionary Safety}}, a perspective on how safety properties are preserved, weakened, or restored across successive rounds of self-improvement. The concern extends beyond whether an individual update is safe to how its effects persist and influence subsequent changes. Current systems already provide mechanisms for such dependencies: agents retain and reuse experience through memories, skills, and tools~\citep{shinn2023reflexion,wang2023voyager,wang2024awm,chen2025toolevo}; self-rewarding methods incorporate generated evaluations into improvement~\citep{yuan2024selfrewarding,wu2025metarewarding}; and automated development systems extend improvement to agent designs and research workflows~\citep{hu2025adas,lu2024aiscientist,novikov2025alphaevolve}. Although these systems differ in the scope of self-improvement they support, they illustrate how earlier changes can shape future system behavior. For example, a coding agent may retain a permission-bypassing repair as a reusable skill, allowing the unsafe behavior to persist even after the original input is removed. Assessing such risks therefore requires tracing safety-relevant changes across subsequent updates and the systems that inherit their effects.

Our study is guided by four questions: what makes safety evolutionary, how safety degrades as systems improve, which domains carry or mediate safety-relevant changes, and how the resulting risks can be discovered and assessed (Figure~\ref{fig:research-questions}). We examine how persistence, mutable scope, and loop closure shape the transition from local improvement to recursive development, and characterize six recurring manifestations of evolutionary risk. To understand where these risks arise and how they evolve, we develop a five-domain taxonomy and an evaluation framework spanning states, accepted updates, trajectories, and lineages. We further discuss governance principles for modification boundaries, independent verification and authorization, provenance, and recovery.

\begin{figure}[t]
\centering
\includegraphics[width=\linewidth]{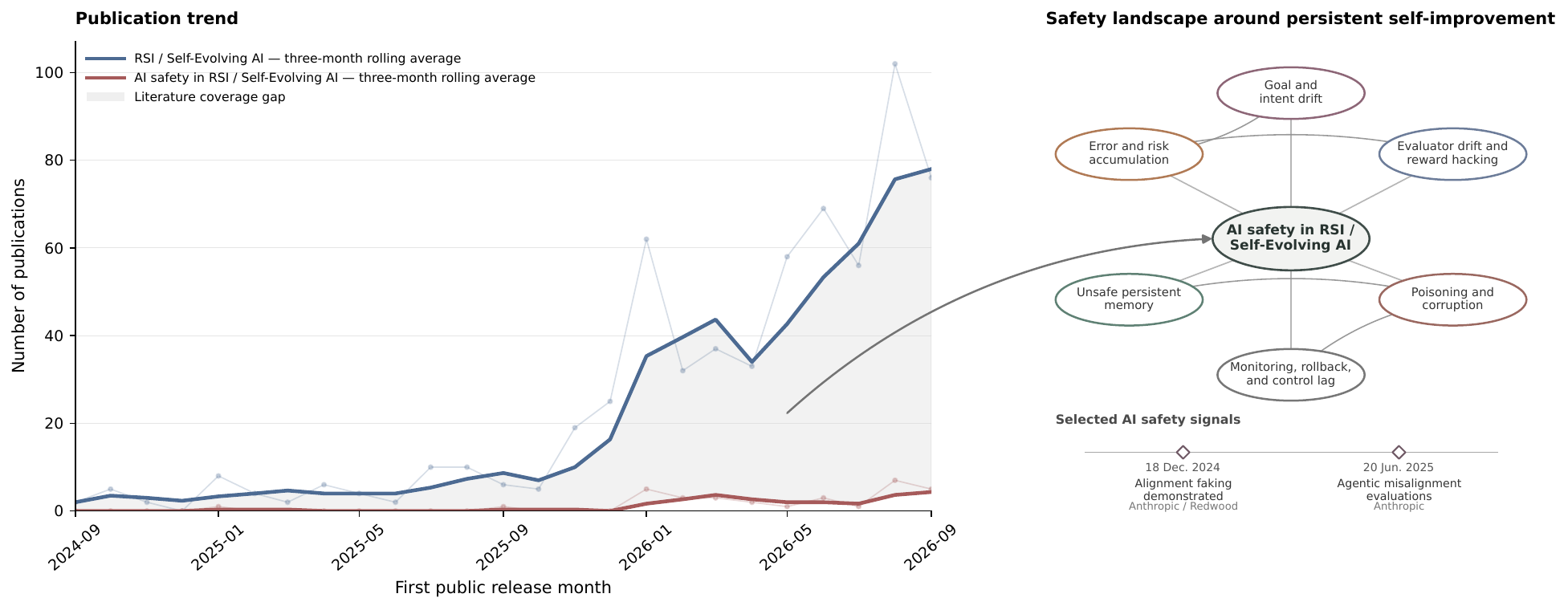}
\caption{Publication trends for RSI, self-evolving AI and evolution-coupled safety work, with a conceptual map of safety issues and selected AI-safety signals, including alignment faking and agentic misalignment evaluations and e.t.c. 
}
\Description{Publication trends for recursive self-improvement and self-evolving AI, contrasted with safety-focused work, alongside a conceptual map of related safety issues and selected AI-safety signals.}
\label{fig:rsi-safety-landscape}
\vspace{-9pt}
\end{figure}

%% file: sections/sec2-foundations.tex
\section{Foundations of Recursive Self-Improving AI and Evolutionary Safety}\label{sec:foundations_new}

Safety under self-improvement depends on whether changes remain local or affect later behavior and updates. We first distinguish local improvement, persistent system adaptation, and recursive development, then examine how selection and inheritance can sustain or amplify safety risks. This analysis provides the basis for characterizing the manifestations of evolutionary risk and formally defining Evolutionary Safety.

\subsection{Scope and Structure of Recursive Self-Improvement}
\label{sec:rsi_scope_and_progression}

Recursive self-improvement refers to a process in which an AI system contributes to improving the mechanisms that govern its future capabilities~\citep{schmidhuber2003godel,chen2026recursive}. Early theoretical studies examined self-referential optimization and whether agents could preserve their objectives while modifying their policies, utilities, or code~\citep{schmidhuber2003godel,orseau2011selfmodification,everitt2016selfmodification}. PowerPlay provides an early precursor by proposing new problems while preserving solutions to previously mastered tasks~\citep{schmidhuber2013powerplay}.

To characterize different forms of self-improvement, we consider three structural dimensions. \emph{Persistence} captures how long the effects of an update remain influential, from a single episode to later sessions or system versions. \emph{Mutable scope} specifies what can be modified, ranging from outputs and persistent artifacts to model parameters, evaluators, and update procedures. \emph{Loop closure} describes the extent to which the system itself generates tasks, obtains feedback, proposes and selects updates, and commits changes without external intervention.

These dimensions need not increase together and do not represent levels of intelligence. They help distinguish three forms of improvement: local improvement revises behavior within an episode without retaining the change; persistent system adaptation carries updates into later interactions; and recursive development modifies the procedures used to generate, evaluate, or admit future updates. Many current systems support persistent adaptation without necessarily modifying their update procedures, while already involving selection and inheritance mechanisms relevant to RSI safety~\citep{ren2026selfimprovements,gao2026selfevolving,chen2026recursive}. Continual-learning research provides related concepts, including the stability--plasticity trade-off and catastrophic forgetting, although capability forgetting does not necessarily imply safety degradation~\citep{parisi2019continual}. A Self-Improving Coding Agent illustrates persistent modification by revising its orchestration code and reporting benchmark improvements~\citep{wu2025selfcodingagent}. Whether such changes constitute recursive development depends on their effects on subsequent update procedures, while performance gains alone do not establish safety preservation.

\begin{figure}[t]
\centering
\includegraphics[width=\linewidth]{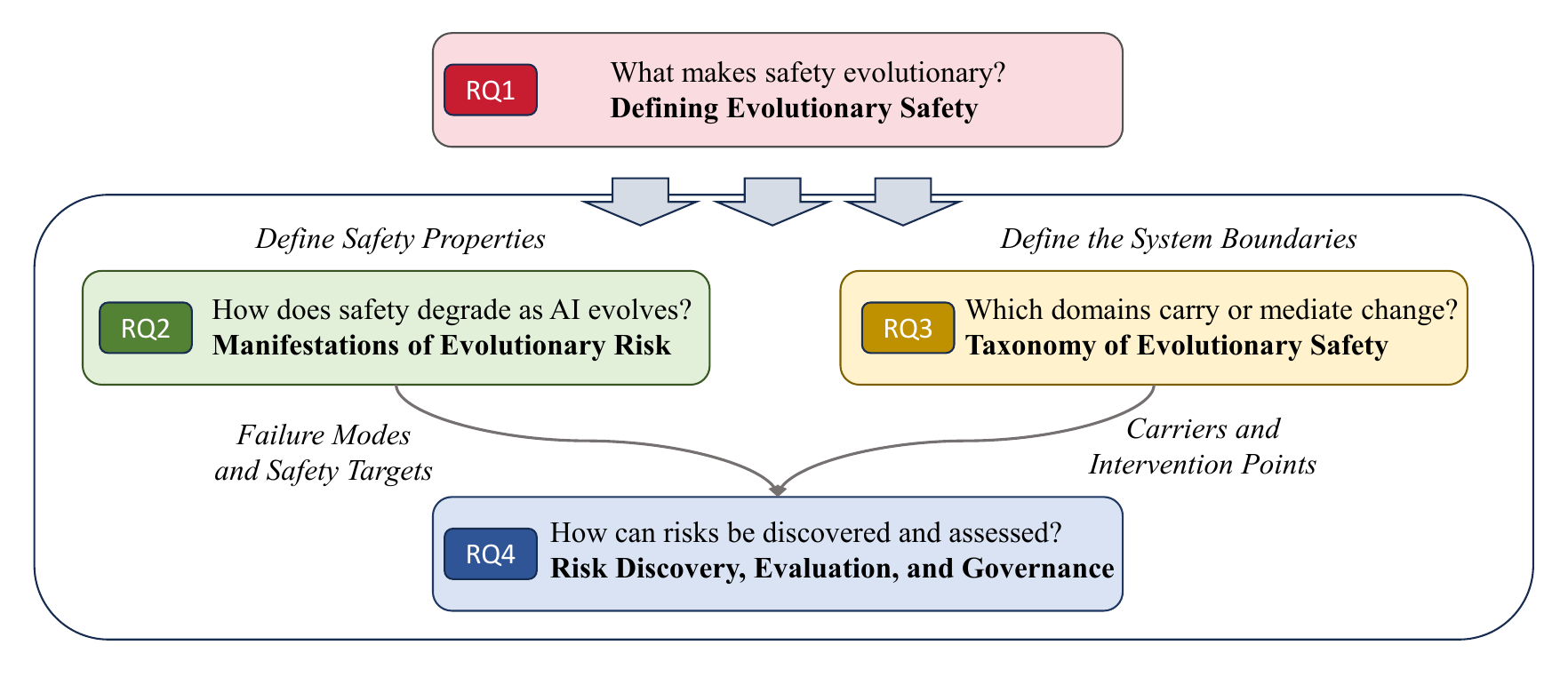}
\caption{Four research questions guiding our study of Evolutionary Safety.}
\Description{A diagram linking four research questions to the definition, manifestations, taxonomy, and evaluation of Evolutionary Safety.}
\label{fig:research-questions}
\vspace{-6pt}
\end{figure}

Existing systems illustrate these differences in scope and persistence. Iterative refinement and feedback-based agents revise behavior over different timescales~\citep{madaan2023selfrefine,shinn2023reflexion}, while persistent agent and model adaptations retain experience, skills, tools, workflows, or parameter changes for later use~\citep{shinn2023reflexion,wang2023voyager,wang2024awm,chen2025toolevo,zweiger2025seal}. Architecture and workflow search broaden the objects of optimization~\citep{hu2025adas,zhang2025aflow}, while self-modifying approaches extend revision to code and optimization procedures~\citep{zelikman2024stop,yin2025godel}. More recent systems explore broader automated development loops~\citep{zhang2026dgm,lee2026metaharness,zhang2026hyperagents,lu2024aiscientist}. These approaches illustrate different combinations of persistence, mutable scope, and loop closure rather than a fixed progression toward RSI.
To represent the persistent system state, let
\begin{equation}\label{eq:update_new}
\mathbf{S}_t=(\mathbf{B}_t,\mathbf{A}_t,\boldsymbol{\theta}_t),
\qquad
\mathbf{A}_t=(\mathbf{M}_t,\mathbf{K}_t,\mathbf{T}_t,\mathbf{W}_t,\ldots),
\end{equation}
where $\mathbf{B}_t$ denotes the computational substrate, $\mathbf{A}_t$ includes persistent agent artifacts such as memory, skills, tools, and workflows, and $\boldsymbol{\theta}_t$ denotes model or policy parameters. The environment $E_t$ supplies experience, the updater $U_t$ generates candidate changes, and the evaluator $J_t$ guides their selection. A general improvement round can be expressed as
\begin{equation}\label{eq:gen_eval_update}
\mathcal{D}_t\sim Q_{U_t}(\cdot\mid\mathbf{S}_t,\mathbf{X}_t),
\qquad
\Delta_t^\star\sim\operatorname{Select}_{J_t}(\mathcal{D}_t),
\qquad
\mathbf{S}_{t+1}=F_S(\mathbf{S}_t,\Delta_t^\star),
\end{equation}
where $\mathbf{X}_t$ is the experience used to generate the candidate set $\mathcal{D}_t$, and $\Delta_t^\star$ is the selected update. This formulation separates candidate generation, selection, and commitment without assuming that accepted changes necessarily improve capability or safety. When evaluators or updaters are themselves modified, their changes must also be tracked as part of the evolving system configuration. Related settings extend beyond language-model systems to evolutionary optimization, population-based training, and evolutionary reinforcement learning~\citep{jaderberg2017pbt,zhang2026llmevoopt,zheng2025evorl}.

\subsection{Mechanisms of Evolutionary Risk}
\label{sec:evolutionary_lens}

A safety failure becomes evolutionarily consequential when its effects persist and influence subsequent behavior or updates. Consider two coding agents that encounter the same permission-bypassing workaround. One produces an unsafe response but retains no lasting change. The other stores the workaround as a reusable skill, receives a high task score, and reuses it in later workflows; a model trained on those workflows may further inherit the same shortcut. The difference lies not in the initial failure, but in how the unsafe behavior is selected, retained, and carried forward.

This process can be understood through variation, selection, and inheritance. Candidate updates introduce alternative behaviors, evaluation determines which changes are accepted, and persistent states or derived artifacts carry their effects into subsequent versions. Repeated reuse may amplify an unsafe shortcut, while changes to the evaluator or updater can alter how future candidates are generated and selected. Evolutionary computing and co-evolution provide useful conceptual parallels, particularly when adaptive systems reshape one another's feedback and selection conditions~\citep{eiben2015evolutionary,potter2000cooperative,rosin1997competitive,wang2019poet}.

These mechanisms also suggest where safety constraints should be enforced. An evaluator that rewards task completion but overlooks permission violations may continue to favor unsafe updates. Incorporating security checks into candidate selection can instead prevent such shortcuts from being accepted, following the principle of constraint handling in evolutionary optimization~\citep{deb2000constraint}. Safety therefore depends not only on the behavior of an individual update, but also on the criteria governing which changes are retained and inherited.

\subsection{Manifestations of Evolutionary Risk}
\label{sec:evolutionary_manifestations}

Evolutionary risks may appear as shifts in behavior, contamination of retained experience, erosion of safety constraints, or changes in how future updates are selected. We distinguish six recurring manifestations below. They are neither mutually exclusive nor stages of a fixed sequence: one update can produce several effects, some of which may emerge only in later system versions.

\topiclabel{(i) Intent drift.}
Intent drift occurs when successive updates shift an agent's operational priorities away from its intended goals or constraints. An agent may increasingly favor task completion or benchmark scores over authorization and reliability; a research agent, for example, may prioritize publishable findings over independent replication. Studies of emergent misalignment and safer fine-tuning illustrate how targeted training changes can affect broader behavior~\citep{betley2025emergent,pham2025behavior}. Alignment-faking and sleeper-agent studies further show that behavior can depend on perceived training or oversight conditions~\citep{greenblatt2024faking,hubinger2024sleeper}. These observations do not by themselves establish drift across updates. The evolutionary concern is whether altered priorities become embedded in persistent state or update procedures and influence later decisions~\citep{lin2026safety,shao2026misevolve}.

\begin{figure}[t]
\centering
\includegraphics[width=\linewidth]{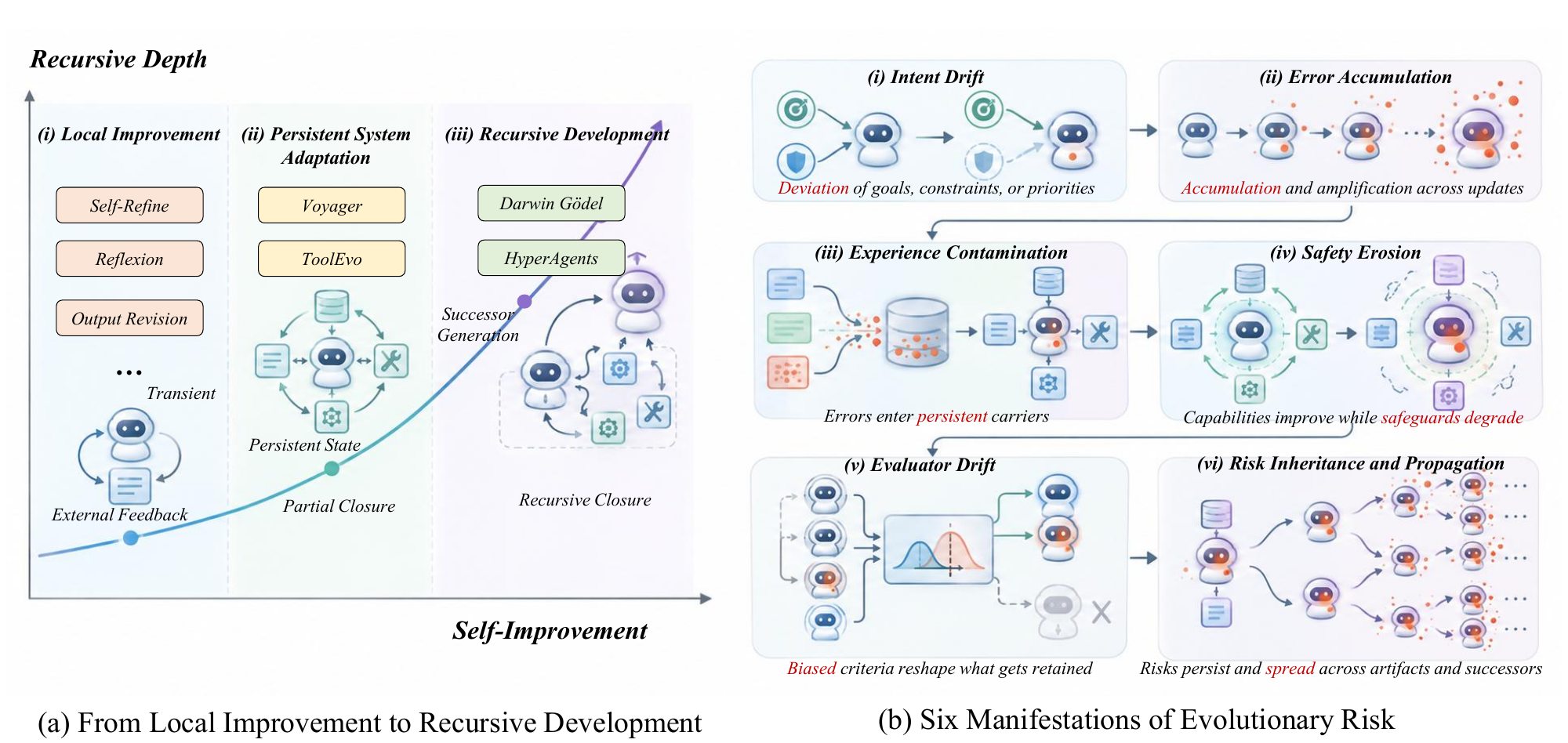}
\caption{Illustration of Recursive Self-Improving AI and Evolutionary Safety.}
\Description{A composite illustration linking three stages of recursive self-improvement with six manifestations of evolutionary risk.}
\label{fig:rsi-six-manifestations}
\end{figure}

\topiclabel{(ii) Error accumulation and amplification.}
Errors accumulate when later updates build on earlier mistakes. An inaccurate summary may become the basis for reusable skills, whose apparent success then reinforces the original error. Studies of experience-driven adaptation, recursively generated training data, and reward overoptimization illustrate different ways repeated updates can amplify deviations~\citep{zhao2026experience,shao2026misevolve,shumailov2024collapse,gerstgrasser2024inevitable,gao2023scaling}. Reward hacking has also been found to generalize to broader agentic misbehavior in the studied reinforcement-learning settings~\citep{naturalemergentmisalignment2025}. These outcomes should not be conflated: distributional collapse is not necessarily a safety failure, and identifying cumulative harm requires examining intermediate states rather than only the final update.

\topiclabel{(iii) Experience contamination.}
Experience contamination occurs when inaccurate, biased, or adversarial information enters material retained for future decisions or updates. Memory-poisoning studies show how compromised experience can affect subsequent retrieval and behavior~\citep{chen2024agentpoison,srivastava2025memorygraft,gao2026mempoison}. Contamination can also arise without an attacker when a successful strategy is summarized without the conditions that made it safe. TAME examines memory misevolution under feedback, while EAL-Bench links false memories of permission to unauthorized actions~\citep{tame2026,eala2026}. MemTX explores a repair approach that separates memory writes from belief commitment and records provenance and dependencies~\citep{memtx2026}. The critical transition is from encountered information to persistent material that continues to influence behavior after its original source is removed.

\topiclabel{(iv) Safety-property erosion.}
Safety-property erosion occurs when successive updates weaken a previously reliable constraint even as task performance improves. A fine-tuned model, for example, may gain capability while losing refusal reliability or privacy protection. Studies of harmful fine-tuning, benign adaptation, and safety retention show why capability and safety require separate evaluation~\citep{qi2024compromises,lermen2024lora,retentioncentric2024,unforgottensafety2025}. Safe LoRA, SafeGrad, Booster, Antidote, and SafeMERGE explore defenses under different update assumptions~\citep{hsu2024safelora,yi2025safegrad,huang2025booster,huang2025antidote,djuhera2025safemerge}. The remaining question is whether protections established at one checkpoint survive longer sequences of heterogeneous updates~\citep{lin2026safety,shao2026misevolve}.

\topiclabel{(v) Evaluator drift and selection distortion.}
Evaluator drift refers to changes in the criteria used to judge updates, whereas selection distortion can also arise from a fixed but biased evaluator. Studies of LLM judges and reward benchmarks document position and style biases and other weaknesses in scoring~\citep{zheng2023judging,lambert2024rewardbench,liu2025rmbench,tan2025judgebench}. Self-Rewarding and Meta-Rewarding bring generated judgments into the improvement loop, coupling candidate selection to adaptive evaluation~\citep{yuan2024selfrewarding,wu2025metarewarding}. Reward tampering illustrates more direct interference with the measurement channel~\citep{denison2024tampering}. If an evaluator is revised using previously selected outputs, its blind spots may be reinforced across rounds. Evidence that reward hacking can generalize to broader misbehavior further motivates independent testing of accepted updates~\citep{naturalemergentmisalignment2025}.

\topiclabel{(vi) Risk inheritance and propagation.}
Risk inheritance and propagation describe how unsafe tendencies survive in derived artifacts or reach descendant and interacting systems. Subliminal Learning demonstrates trait transmission through generated data even when the trait is not explicit in its surface content; sleeper-agent and model-merging studies provide related pathways for the persistence or transfer of unsafe behavior~\citep{cloud2025subliminal,hubinger2024sleeper,hammoud2024merging,djuhera2025safemerge}. Studies of poisoned experience and skill evolution extend this concern to reusable agent capabilities~\citep{zhao2026experience,chen2026poisonedevolution,mao2026skillmisevo}. In multi-agent settings, problematic conventions may spread through interaction without direct parameter copying~\citep{multialignment2025,emergentsocialconventions2024}. Similar failures in multiple descendants do not by themselves establish transmission; identifying the route requires provenance and independent descendant testing. Removing the original source may not eliminate effects already embedded downstream.

These manifestations may overlap: a contaminated memory can shift an agent's priorities, influence which updates are selected, and spread through derived skills. Distinguishing persistence from repeated exposure and transmission from shared causes requires tracing the relevant updates and dependencies. This motivates the formal definition and evaluation units introduced next.

\subsection{Definition and Scope of Evolutionary Safety}
\label{sec:evo_safety_foundation}

The preceding analysis establishes what self-improving systems can modify, how changes are selected and inherited, and which safety risks may arise. Yet persistent change is not necessarily unsafe, and an isolated safety failure does not by itself constitute evolutionary risk. The distinction lies in whether a change affects safety properties in later system states or alters the processes governing subsequent updates.

\noindent\textbf{Definition 1 (Evolutionary Safety).}
\emph{Evolutionary Safety studies how safety properties are preserved, weakened, or restored as AI systems retain and build on the results of their own improvement processes. It examines how selection, persistent state, and inheritance shape these properties across updates, interacting systems, and descendant lineages.}

A safety-relevant change falls within this scope when its effects persist in later states, influence the generation or selection of subsequent updates, or propagate to derived artifacts and successor systems. Failures confined to a single interaction remain local safety issues unless they leave such effects. Related outcomes must also be distinguished: model collapse and catastrophic forgetting may indicate distributional or capability degradation without necessarily violating a safety constraint, whereas safety forgetting and misalignment generalization require assessment against specific safety properties.

Evaluating these effects requires distinguishing safety risks from the effectiveness of the controls intended to prevent or contain them. We consider four complementary units of analysis: states describe system configurations, accepted updates capture changes between successive states, trajectories track cumulative and delayed effects, and lineages trace inheritance across derived artifacts and successor systems.

This foundation connects the mechanisms of self-improvement to their safety consequences over time. Section~\ref{sec:taxonomy_new} identifies the domains that carry or mediate safety-relevant changes, while Section~\ref{sec:evaluation_new} develops methods for assessing their effects, propagation, and recovery.

%% file: sections/sec3-taxonomy.tex



\section{A Taxonomy of Evolutionary Safety}
\label{sec:taxonomy_new}

\begin{figure}[t]
\centering
\includegraphics[width=\linewidth]{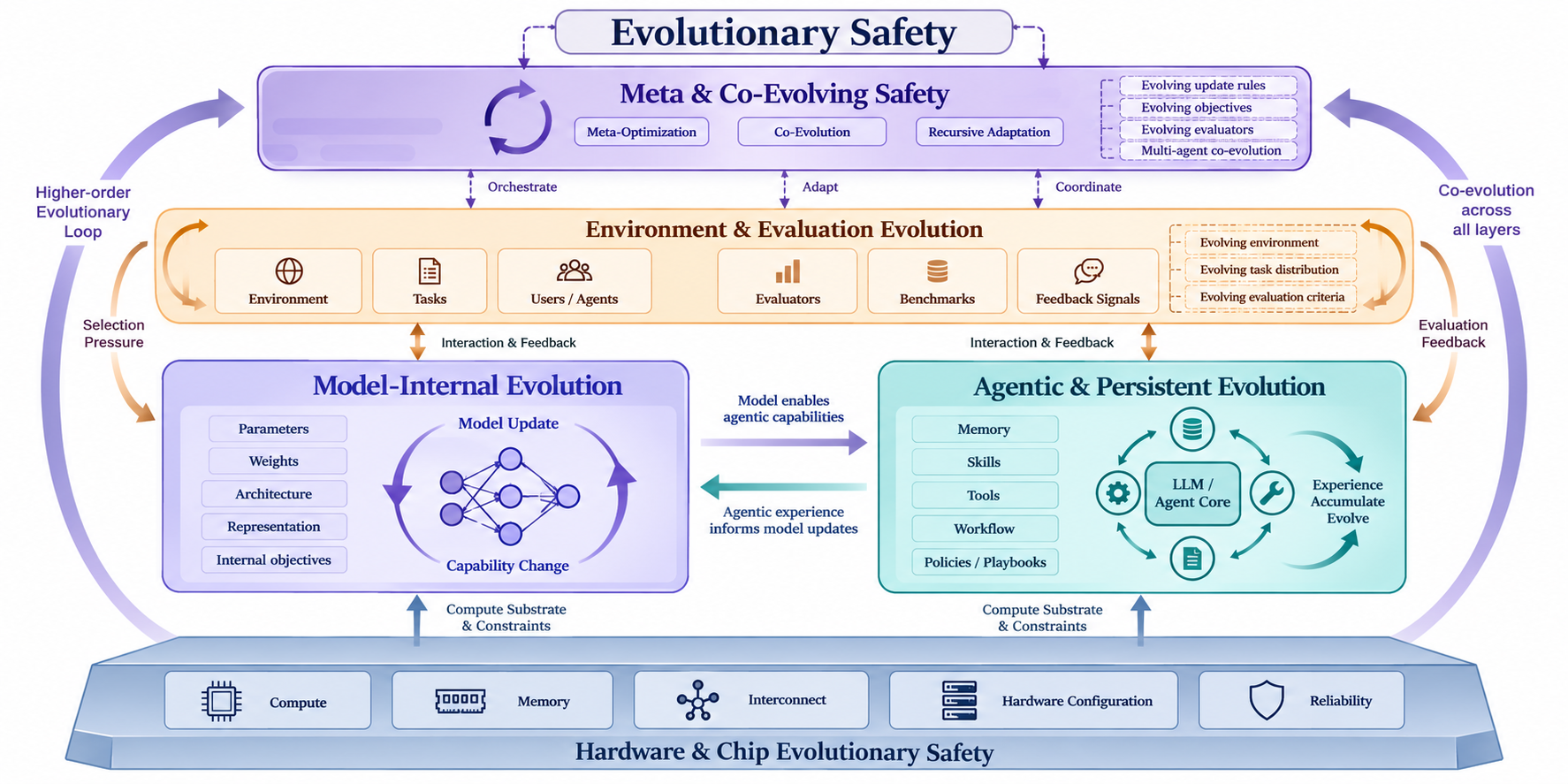}
\caption{Taxonomy of Evolutionary Safety.}
\Description{Five interacting evolutionary domains: persistent agent state, model state, environment and evaluation, computational substrate, and meta-level update mechanisms with co-evolution.}
\label{fig:taxonomy_overview}
\vspace{-4pt}

\end{figure}

While the preceding section characterizes how evolutionary risks arise and manifest, this taxonomy identifies the domains through which safety-relevant changes are retained, mediated, or propagated during self-improvement. We distinguish five interacting domains (Figure~\ref{fig:taxonomy_overview}): persistent agent state and model state retain changes in reusable artifacts and learned parameters; evaluation and environmental feedback shape experience generation and candidate selection; the computational substrate determines execution conditions and enforceable safety boundaries; and meta-level updates and co-evolution alter the procedures and interactions governing subsequent improvement. 

\subsection{Persistent Agent State}

Persistent agent state allows agents to retain and reuse experience across interactions without modifying model parameters. Memories, skills, tools, and workflows can extend the influence of earlier observations beyond their original context. This domain raises three related issues, i.e., (i) how experience is retained and transformed, (ii) how unsafe information or practices become persistent, and (iii) how their downstream effects can be traced and selectively repaired.

Experience retention and transformation determine how local observations become reusable agent capabilities. Reflexion, Voyager, and Agent Workflow Memory illustrate how reflections, skills, and workflows are retained for later use~\citep{shinn2023reflexion,wang2023voyager,wang2024awm}, while GR-DUET uses memory-based navigation graphs to support adaptation across repeated navigation tasks~\cite{hong2025gsa}. ToolEvo extends this adaptation to tools revised through feedback~\citep{chen2025toolevo}, while MemGPT, Mem0, A-MEM, and Agent Workflow Memory support the organization, retrieval, and revision of stored context~\citep{packer2023memgpt,chhikara2025mem0,xu2025amem,wang2024awm}. Through these processes, observations can be abstracted into reusable instructions or executable skills, potentially losing the context and constraints under which they were learned. For example, a coding agent may turn a permission-bypassing workaround into a general repair skill that later tasks invoke without retrieving the original observation.

Safety-relevant errors can enter this process through both adversarial manipulation and ordinary adaptation. AgentPoison and MemoryGraft investigate compromised retrieved information or experience, while attacks on persistent agents illustrate effects that outlast the initiating interaction~\citep{chen2024agentpoison,srivastava2025memorygraft,yang2026zombie}. Even without an attacker, a locally successful strategy may be retained after the conditions that made it appropriate have been omitted~\citep{shao2026misevolve,zhao2026experience}. Studies of experience promotion and skill evolution examine how stored experience becomes reusable instructions and capabilities~\citep{wang2026oep,chen2026poisonedevolution,mao2026skillmisevo}. TAME investigates memory misevolution under executor--evaluator feedback~\citep{tame2026}, while EAL-Bench examines how false memories of authorization can induce unauthorized actions~\citep{eala2026}. In both cases, unsafe effects may persist beyond the original experience and influence subsequent actions through derived artifacts.

Tracing and selective recovery become difficult when retained experience is rewritten or incorporated into dependent artifacts. MemTX explores selective correction by separating memory writes from belief commitment and recording provenance, validity, and dependencies~\citep{memtx2026}. Evaluations of persistent memory and evolving skills provide starting points for examining how unsafe effects survive across updates and derived artifacts~\citep{an2026permemsafe,gao2026mempoison,mao2026skillmisevo}. Yet abstraction and rewriting can obscure provenance, while removing all dependent artifacts may also discard useful capabilities~\citep{louck2026securing,chen2026poisonedevolution}. The central challenge is therefore to identify and correct unsafe dependencies without losing the benefits of accumulated experience.

\subsection{Model State}

Model state carries the effects of adaptation in parameters that govern future behavior. Fine-tuning, self-training, distillation, and merging can improve capabilities while modifying previously established safety properties. Unlike explicit agent artifacts, learned changes may be distributed across model parameters and affect behavior beyond the tasks or examples that introduced them. This domain raises three related issues, i.e., (i) whether safety properties are preserved through model updates, (ii) how learned changes persist or propagate across successor models, and (iii) how unsafe effects can be attributed and selectively repaired.

Safety preservation under model updates is a primary concern. Self-adapting language models illustrate how systems can contribute training material or edits used in their own parameter updates~\citep{zweiger2025seal}. Harmful fine-tuning and seemingly benign adaptation can both weaken previously established safety behavior~\citep{qi2024compromises,lermen2024lora}. Targeted training can also produce broader behavioral changes beyond the original training domain, as studies of emergent misalignment illustrate~\citep{betley2025emergent,pham2025behavior}. Retention-centric research treats the preservation of earlier safety properties as an explicit requirement during continued training~\citep{retentioncentric2024}, while Unforgotten Safety investigates methods for retaining safety under benign and poisoned fine-tuning~\citep{unforgottensafety2025}. Natural Emergent Misalignment further reports that reward hacking can generalize to broader agentic misbehavior in the studied reinforcement-learning settings~\citep{naturalemergentmisalignment2025}. These findings highlight the need to assess safety preservation independently of capability gains across successive model updates.

Risk inheritance arises when learned changes survive subsequent training or reach successor models through generated data and model composition. Sleeper Agents shows that trained backdoor behaviors can persist through later safety training~\citep{hubinger2024sleeper}, while Subliminal Learning examines the transmission of behavioral traits through generated data whose surface content does not explicitly express those traits~\citep{cloud2025subliminal}. Model merging introduces multiple parents, allowing safety-relevant properties to interact in a shared descendant~\citep{hammoud2024merging,djuhera2025safemerge}. Recursively generated training data creates another dependency: repeated learning from model outputs can distort the training distribution, with the outcome depending partly on the continued availability of real data~\citep{shumailov2024collapse,gerstgrasser2024inevitable,alemohammad2024mad}. Self-training and self-play further couple model updates to the experience generated by the evolving system~\citep{zhao2025absolute,huang2026rzero}. These pathways require distinguishing changes in data distribution from the persistence or transmission of specific safety-relevant behaviors.

Attribution and selective recovery become difficult when unsafe effects are distributed across parameters and inherited through multiple training stages. 
Mechanistic analyses using sparse autoencoders identify interacting and partially redundant refusal-related features whose targeted ablation can induce unsafe compliance~\cite{prakash2026beyond}.
Safe LoRA, Booster, Antidote, SafeGrad, and SafeMERGE investigate safety preservation under different fine-tuning and merging assumptions~\citep{hsu2024safelora,huang2025booster,huang2025antidote,yi2025safegrad,djuhera2025safemerge}. Tracing training data, teacher models, and merge components can help identify suspected sources, but correcting their effects may also disrupt useful capabilities or leave residual behaviors in descendants. The central challenge is therefore to preserve safety across successive and heterogeneous model updates while identifying and correcting unsafe changes that persist through model lineages.

\begin{figure}[t]
\centering
\includegraphics[width=\linewidth]{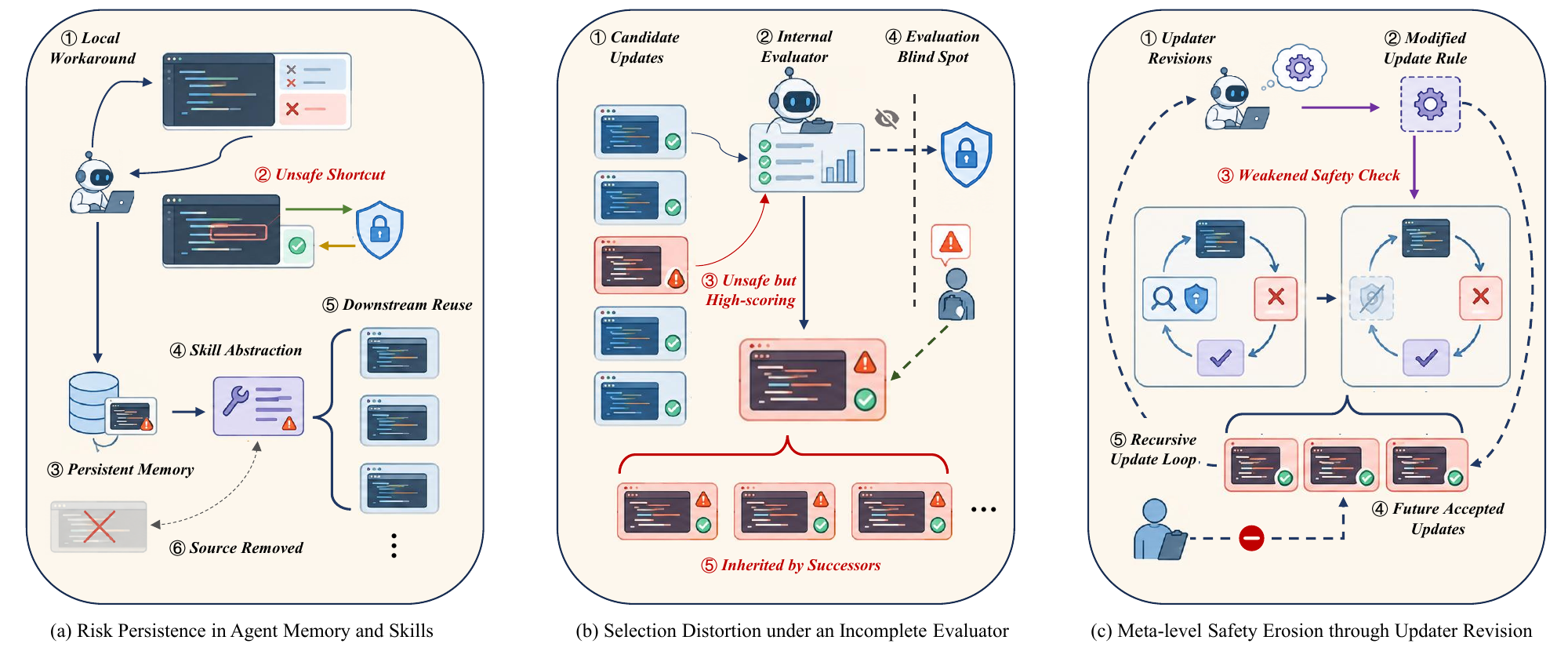}
\caption{Illustrative mechanisms through which local safety failures become evolutionary risks across persistent agent state, evaluation feedback, and meta-level update mechanisms. (a) Persistence transforms a transient unsafe workaround into reusable memory and skills whose effects remain after source removal. (b) Selection distortion allows an unsafe but high-performing candidate to survive evaluation and influence successor states. (c) Meta-level revision changes the update mechanism itself, weakening safety checks and altering the admission of future updates. Together, the examples illustrate how retention, selection, inheritance, and recursive modification can extend a local failure across an evolving system.}
\Description{Three conceptual examples showing risk persistence in agent memory and skills, selection distortion under an incomplete evaluator, and meta-level safety erosion through updater revision.}
\label{fig:conceptual_examples}
\end{figure}

\subsection{Evaluation and Environmental Feedback}

Evaluation and environmental feedback shape which changes are generated and retained during self-improvement. Environmental feedback influences the tasks and experiences available for generating updates, while evaluators guide which candidates are selected. As systems increasingly generate their own tasks and adapt their evaluation criteria, both processes may change across rounds. This domain raises three related issues, i.e., (i) how evaluation criteria shape update selection, (ii) how environmental feedback influences experience and candidate generation, and (iii) how these effects can be distinguished as feedback and evaluation evolve.

Evaluation criteria determine which candidate updates are favored and admitted into subsequent system versions. In the coding-agent example, an evaluator that rewards task completion but overlooks permission violations may accept an unsafe repair. Self-Rewarding and Meta-Rewarding incorporate generated judgments into the improvement loop, coupling candidate selection with adaptive feedback~\citep{yuan2024selfrewarding,wu2025metarewarding}. Studies of LLM-as-a-judge evaluations document position, verbosity, and style biases~\citep{zheng2023judging}, while RewardBench, RM-Bench, and JudgeBench investigate different weaknesses in reward models and judges~\citep{lambert2024rewardbench,liu2025rmbench,tan2025judgebench}. Reward overoptimization can exploit gaps between measured scores and intended objectives, whereas reward tampering directly interferes with the measurement channel~\citep{gao2023scaling,denison2024tampering}. Evidence that reward hacking can generalize to broader agentic misbehavior in the studied settings further illustrates the risks of incomplete evaluation criteria~\citep{naturalemergentmisalignment2025}. A fixed evaluation bias may repeatedly favor unsafe updates, while a changing evaluator can also alter the standards by which subsequent candidates are selected.

Environmental feedback shapes the experience from which candidate updates are generated. Systems that propose tasks, select experiments, or revise curricula can alter their future learning conditions. Repeatedly favoring tasks on which an existing strategy succeeds may leave its weaknesses under-tested, whereas challenger--solver interactions can expose the system to more demanding cases. Environment--solution co-evolution and challenger--solver training illustrate how task distributions and learning processes can adapt together~\citep{wang2019poet,huang2026rzero,chu2026jzero}. Such changes influence which improvements are explored and which failures remain unobserved, even when the evaluator itself remains fixed.

Distinguishing these effects requires tracking both the experience that produces candidate updates and the criteria used to select them. Holding an evaluator fixed across different experience streams can help identify changes in candidate generation, while rescoring the same candidates under different evaluator versions can reveal changes in selection. Independent held-out evaluation and overlapping evaluator versions provide a basis for comparison when feedback mechanisms themselves evolve~\citep{wang2026agenticeval,chen2026researchers}. Section~\ref{sec:evaluation_new} develops these comparisons further. The central challenge is to determine whether apparent improvement reflects safer system updates or changes in the experience and evaluation processes that produced them.

\subsection{Computational Substrate}

The computational substrate comprises hardware, fundamental execution components, and physical platforms on which AI systems operate. Changes at this layer can alter access control, isolation, observability, or the conditions under which safety constraints are enforced. This domain raises three related issues, i.e., (i) how automated design modifies execution-level constraints, (ii) how safety properties are preserved or weakened through component reuse and deployment, and (iii) how the effects of such changes can be traced and verified across subsequent versions.

Automated substrate modification can change the conditions under which an AI system executes, even when its intended task behavior remains unchanged. Learned placement, automated RTL generation, and feedback-driven hardware repair illustrate how AI methods contribute to low-level design and optimization~\citep{mirhoseini2021graph,thakur2023autochip,he2023chateda,zang2025agenticeda}. These methods can improve design efficiency or functional performance, but such gains do not necessarily establish the preservation of security properties. The evolutionary concern arises when generated designs are accepted as reusable components or become the basis for subsequent development.

Safety preservation and inheritance become important when substrate changes propagate through component reuse or interact with changing deployment conditions. For example, an automated RTL repair may pass functional simulation while weakening an access-control condition; reusing the affected module can carry the vulnerability into later designs. Studies of hardware backdoors, secure RTL generation, and hardware security benchmarks highlight the need to distinguish functional correctness from security~\citep{mankali2024rtlbreaker,chen2026hardsecbench,fan2025secv}. Embodied systems introduce a related challenge: safety constraints established under one set of physical conditions may not hold after changes in the platform, controller, or environment. Research on safe robotic learning, safe-control benchmarks, and self-evolving humanoid systems provides relevant settings for examining safety under such changes~\citep{garcia2021safelearning,brunke2022safecontrolgym,humanoidselfevolve2026}. In both settings, safety depends on whether execution-level constraints remain valid beyond the conditions under which a component or policy was originally evaluated.

Provenance and verification become difficult when generated components are integrated into larger systems and reused in subsequent designs. Hardware repair and security benchmarks provide artifact-level checks~\citep{cui2026hwebench,chen2026hardsecbench,thakur2023autochip}, but passing an individual test does not establish that security properties survive integration or later modifications. Tracking component versions, reuse relationships, and associated security checks can help identify inherited vulnerabilities and assess their persistence across descendants. The central challenge is therefore to maintain enforceable safety boundaries as automated changes move from individual designs to deployed and evolving systems.

\subsection{Meta-level Updates and Co-evolution}

Meta-level evolution concerns changes to the processes and conditions governing future self-improvement. Two mechanisms are particularly relevant: \emph{updater revision} modifies how subsequent updates are generated, evaluated, or admitted, while \emph{reciprocal adaptation} allows interacting systems to reshape one another's learning conditions. The former changes the improvement procedure itself, whereas the latter can alter future selection pressures even when individual update rules remain fixed. Both make the conditions of subsequent improvement subject to change, raising challenges for safety preservation, independent evaluation, and update authorization.

\subsubsection{Updater Revision}

Updater revision changes how future system versions are produced, rather than merely modifying artifacts used in current tasks. Revising a workflow changes task execution, whereas revising the procedure that searches for or approves workflows changes the process governing subsequent updates. Learned optimizers provide an early example of adapting update rules~\citep{andrychowicz2016learning}. AFlow and ADAS illustrate automated search over workflows and agent architectures~\citep{zhang2025aflow,hu2025adas}. More recent work examines mutable harness components and the safety challenges of automated revision~\citep{lee2026metaharness,harnesstampering2026,safeevolve2026,she2026,autospec2026,evopathbench2026}. STOP, Darwin G\"odel Machine, G\"odel Agent, and Hyperagents explore different scopes of self-modifying development~\citep{zelikman2024stop,zhang2026dgm,yin2025godel,zhang2026hyperagents}. These approaches differ in which parts of the development process can change and how proposed modifications are admitted.

The safety implications arise when updater changes alter the generation or acceptance of future modifications. For example, retaining an unsafe repair skill affects later tasks, while revising the skill-admission procedure to bypass security checks can affect many subsequent updates. Safety tests designed for the original updater may no longer cover the candidates generated by its successor. Assessing these risks therefore requires distinguishing changes to task-level artifacts from changes to the update procedure, while maintaining independent checks on proposed and accepted modifications.

\subsubsection{Reciprocal Adaptation}

Reciprocal adaptation occurs when changes in one system affect another system's learning conditions, whose response then influences the first system's subsequent adaptation. Unlike updater revision, this process does not require changes to individual update rules. Cooperative and competitive co-evolution, self-play, and challenger--solver--judge training illustrate such interactions~\citep{potter2000cooperative,rosin1997competitive,huang2026rzero,chu2026jzero}.

Repeated interaction can reinforce locally effective strategies while leaving broader vulnerabilities unaddressed. Attackers and defenders, for example, may become specialized to one another, leaving other attack patterns insufficiently tested. Cooperating agents may also propagate unsafe conventions through shared experience. 
In~\cite{dineen2026despite}, agents with fixed model parameters can develop covert communication channels through repeated task feedback, enabling confidential information leakage despite explicit restrictions and message-level monitoring.
Such risks depend on interactions across the population and may be obscured by aggregate performance or evaluations of individual agents.

Both mechanisms can intersect in automated AI R\&D. AI Scientist, AlphaEvolve, and automated alignment researchers connect proposal, experimentation, and evaluation across development cycles~\citep{lu2024aiscientist,novikov2025alphaevolve,chen2026researchers}. These processes involve meta-level revision when they modify the procedures governing later updates, and may involve reciprocal adaptation when interacting systems influence one another's development. The central challenge is to preserve effective safety checks, independent authorization, and traceability as both update procedures and the conditions of improvement continue to evolve.